\documentclass{isprs}

\usepackage[margin=1in]{geometry} % Set page margins to 1 inch
\usepackage{graphicx}             % Required for inserting images
\usepackage[colorlinks=false, pdfborder={0 0 0}]{hyperref} % Clickable links without blue boxes
\usepackage{fontawesome} % for the GitHub icon
\usepackage{booktabs}
\usepackage{float}               % for [H] placement
\usepackage{array}  
\usepackage{subcaption}
\usepackage{cite}
\usepackage[utf8]{inputenc}   % allow UTF-8 characters in your .tex file
\usepackage[T1]{fontenc}      % good font encoding
\usepackage[margin=1in]{geometry}
\usepackage{graphicx}                % for placement (if you still need it)
\usepackage[section]{placeins}       % for             % if you’re using subfigures
\usepackage{caption}
\usepackage{comment}
\usepackage{soul} % for the command \hl
\usepackage{color} % for the command \textcolor
\usepackage{amsmath} % for align environment
\usepackage{hyperref}
\usepackage{multirow}%
\usepackage[dvipsnames]{xcolor}

\usepackage{fancyhdr}
\begin{document}

%\newpage
\title{A Quantitative Evaluation Framework for Explainable AI in Semantic Segmentation}
\date{}

\author{
Reem Hammoud\textsuperscript{1}, 
Abdul karim GIZZINI\textsuperscript{2}, 
Ali J. Ghandour\textsuperscript{3*}
}

\address{
\textsuperscript{1} American University of Beirut, Beirut, Lebanon \\
\textsuperscript{2} SogetiLabs Research and Innovation (part of Capgemini), Issy Les Moulineaux, 92130, France \\
\textsuperscript{3} National Center for Remote Sensing, CNRS-L, Beirut, Lebanon  \\
\textsuperscript{*} Corresponding Author: aghandour@cnrs.edu.lb
}

% KAO: Remove extra spacing
% Anonymous submissions, authors' names should not be visible
%\author{***** (for review, names must be rendered anonymous)}

% KAO: Remove extra newline
% Anonymous submissions, authors' affiliations should not be visible
%\address{**** (for review, affiliations must be rendered anonymous)}

% If the corresponding author is NOT the final author, always add a % space before the subsequent comma, i.e.
% first author name\textsuperscript{a,}\thanks{Corresponding author} , % second author name \textsuperscript{b}, etc.
% thanks to Niclas Borlin 05-05-2016

%\commission{XX, }{YY} %This field is optional. If filled, XX and YY should be replaced by adequate numbers. See https://www2.isprs.org/commissions/
%\workinggroup{XX/YY} %This field is optional.
\icwg{}   %This field is optional.

\abstract{Ensuring transparency and trust in artificial intelligence (AI) models is essential as they are increasingly deployed in safety-critical and high-stakes domains. Explainable AI (XAI) has emerged as a promising approach to address this challenge; however, the rigorous evaluation of XAI methods remains vital for balancing the trade-offs between model complexity, predictive performance, and interpretability. While substantial progress has been made in evaluating XAI for classification tasks, strategies tailored to semantic segmentation remain limited. Moreover, objectively assessing XAI approaches is difficult, since qualitative visual explanations provide only preliminary insights. Such qualitative methods are inherently subjective and cannot ensure the accuracy or stability of explanations. To address these limitations, this work introduces a comprehensive quantitative evaluation framework for assessing XAI in semantic segmentation, accounting for both spatial and contextual task complexities. The framework systematically integrates pixel-level evaluation strategies with carefully designed metrics to yield fine-grained interpretability insights. Simulation results using recently adapted class activation mapping (CAM)-based XAI schemes demonstrate the efficiency, robustness, and reliability of the proposed methodology. These findings advance the development of transparent, trustworthy, and accountable semantic segmentation models.}

% Keywords

\maketitle

\section{Introduction} \label{sec1}

The increasing use of complex Artificial intelligence (AI) models has highlighted a major issue that lies in the interpretability and transparency of these models. These powerful models often function as black boxes, which means that we cannot see how they make decisions~\cite{fraternali2023black}. This lack of transparency makes it hard to trust them, limits our ability to find and fix errors, and even raises ethical concerns~\cite{kumar2024human}. To address this issue, explainable artificial intelligence (XAI) has emerged~\cite{liu2024human}. XAI aims to make AI models more transparent by offering ways to explain their predictions and behavior. This involves creating methods that show why a model made a specific decision or how it reached a particular outcome. The ultimate goal is to open the black box and provide insights people can easily understand, leading to greater trust and effective use of AI technologies.
However, simply generating explanations is not enough. The effectiveness of XAI methods itself must be ensured and evaluated~\cite{stassin2023experimental}. This evaluation process is crucial for ensuring that explanations are not only understandable but also accurate representations of how the model works internally. Although various evaluation strategies have been proposed for XAI methods in image classification tasks~\cite{lopes2022xai}, the XAI evaluation for semantic segmentation still lacks investigation~\cite{gipivskis2024explainable}. Semantic segmentation involves dividing an image into meaningful regions, which is important for many applications such as object detection and scene understanding~\cite{manakitsa2024review}. The explanations of the segmentation model should consider the complex relationships between pixels and provide information about spatial locations. Additionally, the value of an explanation depends not only on its correctness, but also on how easily it can be understood by the intended audience.

This paper aims to fill this gap by proposing a comprehensive framework to evaluate XAI methods specifically designed for semantic segmentation models. Our approach takes into account the unique constraints that segmentation tasks present, such as spatial coherence, contextual interdependence between regions, and the necessity for explanations that are consistent with human visual perception. In this work, we propose a robust evaluation framework that incorporates four alternative evaluation strategies, each associated with relevant metrics to assess the effectiveness of XAI methods in the context of semantic segmentation. In summary, our contributions can be summarized as follows:

\begin{itemize}
    \item Propose a novel XAI evaluation framework for semantic segmentation that includes comprehensive evaluation strategies and metrics.

    \item Introduce a comprehensive and reliable XAI evaluation strategy that uses the predicted masks and the ground-truth labels. This better fits the constraints of semantic segmentation applications.
    
    % \item Design a dual evaluation approach, pixel-level and image-level assessments. This fusion enhances the overall interpretability of semantic segmentation by ensuring that only the critical features controlling the model's decisions are effectively highlighted.
    
    \item Evaluate the performance of the recently adapted XAI methods for semantic segmentation, ensuring a comprehensive, reliable, and robust XAI evaluation.
\end{itemize}

% The remainder of this paper is organized as follows: Section~\ref{sec2} presents an overview of the XAI evaluation methods. Section~\ref{sec3} illustrates the proposed XAI evaluation framework as well as the foundation of the proposed pixel-based and image-based evaluation methodologies. In Section~\ref{sec4}, the performance evaluation of recently adapted XAI methods for semantic segmentation is performed. Finally, Section~\ref{sec5} concludes the manuscript.

\section{XAI Evaluation: Overview}
\label{sec2}

Recently, researchers have developed various XAI methodologies and metrics to assess explanation methods, particularly for image classification tasks~\cite{fresz2024classification}. 
%
% This section provides an overview of some key evaluation approaches used in previous studies, which form the basis for our proposed XAI evaluation framework for semantic segmentation.
%
The research has primarily focused on the use of objective metrics to evaluate the effectiveness of explanation methods. An example is the evaluation of  Gradient-weighted Class Activation Mapping (Grad-CAM++)~\cite{gcpp} using metrics such as the average drop percentage, percentage increase in confidence and win percentage. These metrics quantify how well the generated explanation maps highlight the relevant regions in the image that contribute to the model's decision-making process. Compared to its predecessor, Grad-CAM~\cite{gc}, Grad-CAM++ demonstrated a lower average drop percentage $36.84\%$ and a higher percentage increase in confidence $17.05\%$, indicating its superior ability to generate accurate explanations. In addition to objective metrics, subjective evaluations that involve human judgments are used to assess the interpretability and trustworthiness of explanation methods. For example, Grad-CAM++ underwent a subjective evaluation in which human subjects compared explanation maps from Grad-CAM and Grad-CAM++, and it was found that Grad-CAM++ invoked greater trust. This approach helps determine whether the explanations align with human intuition and understanding, thereby assessing their practical usefulness in real-world scenarios. Similarly, Score-CAM~\cite{sc} was assessed using metrics such as average drop and average increase in confidence, showing significant improvements over previous methods. These objective evaluations provide a quantitative measure of an explanation method's performance by analyzing how well the explanation maps align with the model predictions.

Evaluations also extend to measuring the localization capability of explanation methods. Grad-CAM and Score-CAM, for example, were evaluated on the basis of their ability to accurately locate objects within images. Grad-CAM’s evaluation involved weakly-supervised localization and segmentation tasks, as well as a pointing game evaluation to measure discriminativeness~\cite{gcpp}. Score-CAM used energy-based localization evaluations~\cite{sc}, demonstrating superior performance in accurately localizing target objects. These evaluations are crucial in tasks where precise identification of object locations is essential. In addition, qualitative evaluations are used through visual inspections and sanity checks to validate the generated explanations. For example, Score-CAM included a sanity check by comparing its outputs against those from a randomly initialized network to ensure that the explanations reflect the model's learned parameters and not random artifacts.

In summary, the evaluation of the XAI methods has focused mainly on classification tasks using a combination of: \textit{(i)} Objective metrics, \textit{(ii)} Subjective evaluations involving human judgments, \textit{(iii)} Localization assessments, and \textit{(iv)} Qualitative validations through visual inspections and sanity checks.
These methodologies provide a comprehensive framework for assessing the accuracy, interpretability, and reliability of explanation methods. Building upon these foundations, our research extends these evaluation strategies to the field of semantic segmentation, proposing customized measures that consider the complexities of segmenting intricate visual scenes.

\section{Proposed XAI Evaluation Framework} \label{sec3}

The question of how different XAI methods shed light on the influential features that influence the pre-trained model's decisions. Through this surge, we propose four XAI evaluation strategies, explained as follows:

\begin{itemize}
    \item \textbf{S1 - Background Only:} 
Removing highlighted pixels from the original image to observe the impact on the model's segmentation score, thus evaluating the crucial importance of the identified features. We \textbf{mask the relevant highlighted pixels} from the original image, creating a new version of the image without these pixels. We fed this masked image for a second time into the model, measuring by how much the segmentation score drops. A big drop means that the highlighted pixels were crucial, showing that the XAI method effectively identified important features.

\item \textbf{S2 - Highlighted Only: } 
Isolating XAI-highlighted pixels from the background to pinpoint the features influencing the model’s decisions. We \textbf{zoom in on the highlighted pixels} from the XAI heat map by masking the background. This brings out a new image of just the highlighted pixels, which will help to better understand which features the model is relying on, without the distraction of other background information. In this case, in contrast to S1, a lower drop in the segmentation score means that the corresponding XAI method is better.

\item \textbf{S3 - Predicted Mask, Ground Truth (PMGT): } Observing the change in the segmentation score using S1 and S2 is not sufficient due to the spatial correlation between neighboring pixels. An XAI method for the segmentation task is supposed to highlight relevant pixels, regardless of whether these relevant pixels belong to the target class or any other classes. Therefore, simply monitoring the change in the segmentation score of the target class is insufficient to track how well the XAI method performs. To resolve this issue, we introduce the PMGT evaluation strategy that can be expressed in two forms as follows:

\begin{itemize}
    \item \textbf{XAI-PM}: where we combine the XAI heatmap with the predicted target areas of the model to see which XAI method highlights the most relevant pixels for model prediction. This allows us to visualize which XAI method shows the most important pixels that help the model make its predictions, even if those predictions are not perfect.

    \item \textbf{XAI-GT}: comparing XAI-highlighted pixels with actual GT data to assess which method accurately identifies pixels that truly contribute to optimizing the model performance. This helps to know which XAI method highlights the pixels that actually help to improve the prediction of the model. The XAI heatmap is wrapped over the ground truth (GT) target mask. The edited image displays both the real target class pixels from the GT mask and the ones that have been highlighted. In this manner, we are able to measure how selective each XAI method is in highlighting those specific pixels that actually would contribute to optimizing the model's performance.
    
\end{itemize}
\end{itemize}

\begin{figure*}[t]
    \centering
\includegraphics[width=1\linewidth]{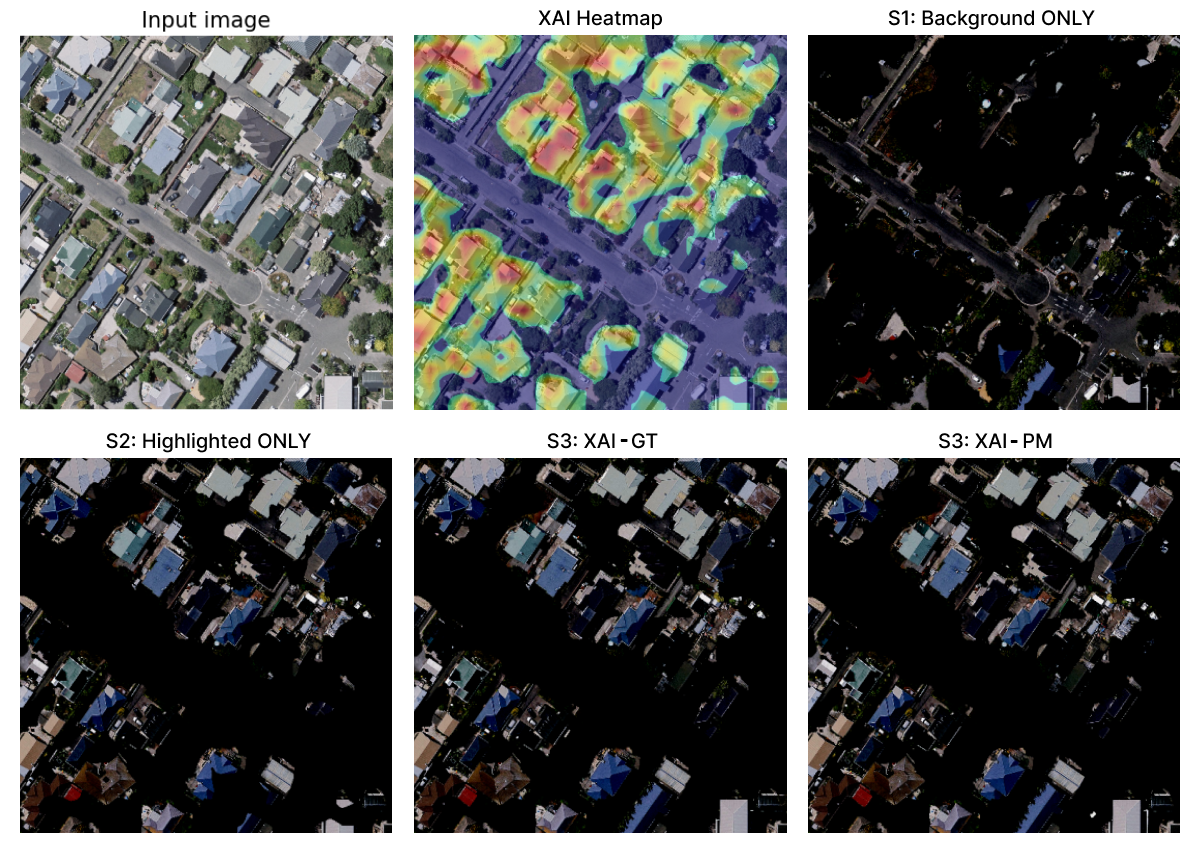}
    \caption{Proposed XAI Evaluation Strategies.}
    \label{fig:propsed_str}
\end{figure*}

To quantitatively assess these strategies, we need to go deeper than simply understanding the final segmentation. Our goal is to evaluate how well these XAI techniques illuminate the crucial features influencing the model's decisions. The objective is to measure how much the XAI highlighted pixels impact each metric. To achieve this, we propose the following pixel-level evaluation metrics:

\begin{itemize}

  \item True Positives (TP): These pixels represent the foundation for successful segmentation. They signify pixels correctly identified by the model and the XAI method as belonging to a specific class.
  
  \item False Negatives (FN): These pixels represent missed opportunities. The model and XAI method fail to identify pixels that genuinely belong to a particular class.
  
  \item False Positives (FP): These pixels represent misleading explanations. The model and XAI method incorrectly assign a class label to pixels that don't belong to that class.
  
\end{itemize}

By analyzing these pixel-level measures, we can calculate essential metrics to assess the effectiveness of the XAI method in terms of:

\begin{itemize}

  \item Precision: This metric reflects the ability of the XAI method to pinpoint relevant features. High precision indicates that the highlighted features genuinely contribute to the model's classification. The precision can be expressed as follows:
  \begin{equation}
  \text{Pre} = \frac{TP}{TP + FP}.
  \end{equation}
  
  \item Recall: This metric signifies the completeness of the XAI method in capturing all the crucial features. A high recall suggests that the XAI method effectively identifies most of the influential features. The Recall can be expressed as follows:

  \begin{equation}
  \text{Re} = \frac{TP}{TP + FN}.
  \end{equation}
  
  \item F1 Score: This metric strikes a balance between precision and recall, providing a more comprehensive picture of the performance of the XAI method. A high F1 score indicates that the XAI method is accurate and comprehensive in highlighting key features. The F1 score can be expressed as follows:

  \begin{equation}
  \text{F1} = 2~ \frac{Pre\times Re}{Pre + Re}.
  \end{equation}

  \item IoU: This metric assesses how well the XAI method's highlighted pixels overlap with the Ground-Truth (GT) mask, evaluating the spatial accuracy of the explanations. Higher IoU scores indicate a better overlap between the XAI method's highlighted regions and the ground truth, reflecting a better localization accuracy. The IOU score can be expressed as follows:

\begin{equation}
  \text{IOU} = \frac{TP}{TP + FP + FN}.
  \end{equation}
    \end{itemize}

\section{Simulation Results} \label{sec4}

To evaluate our proposed XAI evaluation framework, we performed experiments using various adapted CAM-based XAI methods for semantic segmentation~\cite{gizzini2024extending}, including:  Grad-CAM, Grad-CAM++, XGrad-CAM, Score-CAM, Ablation-CAM, and Eigen-CAM.
The experiments are performed on the WHU Building Dataset~\cite{luo2023diverse}, a high-quality resource for building segmentation tasks to test these methods in a real-world context. The simulations are conducted on a GPU-powered environment using CUDA 12, ensuring fast and efficient processing.

\begin{table*}[t]
\centering
\resizebox{\textwidth}{!}{
\begin{tabular}{|c|c|c|c|c|c|c|c|c|}
\hline
Threshold            & XAI/Matrix              & Model & Grad-CAM & Grad-CAM++ & XGrad-CAM  & Score-CAM       & Eigen-CAM      & Ablation-CAM   \\ \hline
\multirow{9}{*}{0.4} & TP Pixels               & 49431 & 45259    & 39680      & 45238  & 25906           & 46990          & 45228          \\  
                     & TP Pixels (\%)          & 93.58 & 85.68    & 75.12      & 85.64  & 49.05           & 88.96          & 85.62          \\  
                     & Drop \%(\textbf{Higher better})  &       & 7.90     & 18.46      & 7.94   & \textbf{\color{ForestGreen} 44.53}  & { 4.62}           & 7.96           \\  
                     & FP Pixels               & 2886  & 2806     & 4275       & 2847   & 3846            & 5622           & 3095           \\  
                     & FP Pixels (\%)          & 1.38  & 1.34     & 2.04       & 1.36   & 1.84            & 2.69           & 1.48           \\  
                     & Increase \% (\textbf{Higher better})  &       & { -0.04}     & 0.66      & -0.02   & 0.46           & \textbf{\color{ForestGreen} 1.31} & 0.10          \\  
                     & FN Pixels               & 3390  & 7562     & 13141      & 7583   & 26915           & 5832           & 7593           \\  
                     & FN Pixels (\%)          & 6.42  & 14.32    & 24.88      & 14.36  & 50.95           & 11.04          & 14.38          \\  
                     & Increase \% (\textbf{Lower better})  &       & 7.90    & 18.46     & 7.94  & \textbf{\color{ForestGreen} 44.53} & { 4.62}          & 7.96          \\ \hline
\end{tabular}%
}
\caption{Pixel-level performance evaluation of S1.}
\label{tbl:s1-pl}
\end{table*}

\begin{table*}[t]
\centering
\resizebox{\textwidth}{!}{%
\begin{tabular}{|c|c|c|c|c|c|c|c|c|}
\hline
Thresholds           & Method      & Model & Grad-CAM & Grad-CAM++ & XGrad-CAM & Score-CAM & Eigen-CAM & Ablation-CAM \\ \hline
\multirow{4}{*}{0.4} & IoU (Micro) & 0.89  & 0.81     & 0.69       & 0.81  & \textbf{\color{ForestGreen}0.46}      & 0.80      & 0.81         \\  
                     & Precision   & 0.94  & 0.94     & 0.90       & 0.94  & \textbf{\color{ForestGreen}0.87}      & 0.89      & 0.94         \\  
                     & Recall      & 0.94  & 0.86     & 0.75       & 0.86  & \textbf{\color{ForestGreen}0.49}      & 0.89      & 0.86         \\  
                     & F1          & 0.94  & 0.90     & 0.82       & 0.90  & \textbf{\color{ForestGreen}0.63}      & 0.89      & 0.89         \\ \hline
\end{tabular}%
}
\caption{Pixel-level performance evaluation of S1. Lower metric value signifies better XAI scheme and vice versa.}
\label{tbl:s1-pll}
\end{table*}

In the conducted experiments, we follow a structured approach to evaluate the performance of the considered XAI methods. First, the XAI methods are applied to the input images, producing corresponding heatmaps highlighting the regions the model deemed important to segment the building target class. These heatmaps are then subjected to a thresholding process using four thresholds (0.2, 0.4, 0.6, 0.8) to isolate the relevant areas of interest. For each threshold, the proposed evaluation strategies, that is, S1 (Background Only), S2 (Highlighted Only), and S3 (PMGT strategy), are applied to generate modified versions of the images. Subsequently, pixel-level metrics, including TP, FP, FN, precision, recall, F1 score, IoU, are then computed to assess the performance of each XAI method quantitatively. Finally, we note that, based on the evaluation across various thresholds, 0.4 demonstrates a good trade-off between the considered metrics, as it provides the most meaningful information on the impact of background removal. Therefore, subsequent analysis will focus exclusively on the results obtained at threshold = 0.4.
%
% This systematic process allowed for a comprehensive comparison of the XAI methods in identifying and localizing key features that influence the model’s decisions.
In the next sections, we will discuss the results per evaluation strategy.

\subsection{S1: Background only}

% \hl{FP: Predicated As building but in GT not building. FN: Predicated as not building but in GT building }

Tables~\ref{tbl:s1-pl} and~\ref{tbl:s1-pll} show the performance evaluation at the pixel-level using the S1 strategy employing the considered thresholds. Recall that in the S1 strategy, the objective is to remove the relevant highlighted pixels that contribute to building segmentation and to keep the background pixels. The highlighted pixels are removed each time according to different thresholds. It is worth mentioning that the XAI scheme that achieves a higher drop in the segmented TP pixels outperforms the other XAI schemes. This is because when the highlighted important pixels are removed, the model is supposed to misclassify the TP pixels within the target class. In contrast, a higher increase in the FP and FN pixels reveals that the corresponding XAI scheme is the best because removing the relevant highlighted pixels should increase the erroneously segmented pixels outside the target class (FP pixels) and the correct unsegmented pixels within the target class (FN pixels), respectively.

% \begin{figure}[t]
%     \centering
%     \begin{subfigure}[t]{0.33\textwidth}
%         \centering
%         \includegraphics[height=4cm]{images/S1_TP.png}
%         \caption{TP Pixels}
%     \end{subfigure}%
%     ~ 
%     \begin{subfigure}[t]{0.33\textwidth}
%         \centering
%         \includegraphics[height=4cm]{images/S1_FP.png}
%         \caption{FP Pixels}
%     \end{subfigure}%    
%     ~
%     \begin{subfigure}[t]{0.33\textwidth}
%         \centering
%         \includegraphics[height=4cm]{images/S1_FN.png}
%         \caption{FN Pixels}
%     \end{subfigure}
%     \caption{Pixel-level performance evaluation for S1 strategy.}
%     \label{fig:s1pxl}
% \end{figure}

As we can notice, Score-CAM retains its position as the top performing method for TP pixel drop (44.53\%), followed by Grad-CAM++ (18.46\%). The performance gap between the top and bottom ranked methods widened, highlighting the robustness of Score-CAM. For FP, Eigen-CAM shows the highest increase (1.31\%), closely followed by Score-CAM (0.46\%) and Grad-CAM++ (0.66\%).  FN analysis again highlighted Score-CAM as the most effective method with an increase of (-44.53\%), strengthening its ability to isolate and preserve critical features in masked images. We can clearly notice that Score-CAM achieves the highest values in both Precision and Recall, confirming that the regions it identified were accurate and essential for correct segmentation. A higher Precision indicates that the model was more focused and less likely to wrongly classify background pixels as part of the target class, while a higher Recall means that it still managed to correctly identify most of the true target pixels. Together, these results reinforce that Score-CAM not only highlights the most important features but also does so with minimal noise or distraction, making it the most reliable and effective XAI method.

\begin{table*}[t]
\centering
\resizebox{\textwidth}{!}{%
\begin{tabular}{|c|c|c|c|c|c|c|c|c|}
\hline
Threshold            & XAI/Matrix              & Model & Grad-CAM        & Grad-CAM++     & XGrad-CAM  & Score-CAM       & Eigen-CAM       & Ablation-CAM \\ \hline
\multirow{9}{*}{0.4} & TP pixels               & 49431 & 28948           & 25871          & 27938  & 39505           & 25067           & 27003        \\  
                     & TP Pixels (\%)          & 93.58 & 54.80           & 48.98          & 52.89  & 74.79           & 47.46           & 51.12        \\  
                     & Drop \%(\textbf{Lower better})   &       & 38.78           & 44.6           & 40.69  & \textbf{\color{ForestGreen} 18.79}  & { 46.12}           & 42.46        \\  
                     & FP pixels               & 2886  & 152690          & 123876         & 148112 & 95918           & 84419           & 141878       \\  
                     & FP pixels (\%)          & 1.38  & 72.94           & 59.18          & 70.76  & 45.82           & 40.33           & 67.78        \\  
                     & Increase \% (\textbf{Lower better}) &       & { 71.56}          & 57.80         & 69.38 & 44.44          & \textbf{\color{ForestGreen} 41.71} & 66.40       \\  
                     & FN pixels               & 3390  & 23873           & 26950          & 24883  & 13316           & 27754           & 25818        \\  
                     & FN pixels (\%)          & 6.42  & 45.20           & 51.02          & 47.11  & 25.21           & 52.54           & 48.88        \\  
                     & Increase \% (\textbf{Lower better}) &       & 38.78          & 44.60         & 40.69 & \textbf{\color{ForestGreen} 18.79} & { 58.96}          & 42.46       \\ \hline
\end{tabular}%
}
\caption{Pixel-level performance evaluation of S2.}
\label{tbl:s2-pl}
\end{table*}

\begin{table*}[t]
\centering
\resizebox{\textwidth}{!}{%
\begin{tabular}{|c|c|c|c|c|c|c|c|c|}
\hline
Threshold            & Method      & Model & Grad-CAM & Grad-CAM++ & XGrad-CAM & Score-CAM & Eigen-CAM & Ablation-CAM \\ \hline
\multirow{4}{*}{0.4} & IoU (Micro) & 0.89  & 0.14     & 0.15  & 0.14  & \textbf{\color{ForestGreen}0.27}      & 0.18      & 0.14         \\  
                     & Precision   & 0.94  & 0.16     & 0.17  & 0.16  & \textbf{\color{ForestGreen}0.29}      & 0.23      & 0.16         \\  
                     & Recall      & 0.94  & 0.55     & 0.49  & 0.53  & \textbf{\color{ForestGreen}0.75}      & 0.47      & 0.51         \\  
                     & F1          & 0.94  & 0.25     & 0.26  & 0.24  & \textbf{\color{ForestGreen}0.42}      & 0.31      & 0.24         \\ \hline
\end{tabular}%
}
\caption{Pixel-level performance evaluation of S2. Higher metric value signifies a better XAI scheme and vice versa.}
\label{tbl:s2-pll}
\end{table*}

Since the S1 strategy masks highlighted pixels, it cannot determine whether the remaining background pixels contain meaningful information. To address this limitation, it is beneficial to evaluate performance using the S2 strategy, where this complementary approach isolates highlighted pixels and evaluates their importance without the distraction of background data, providing deeper insights into the efficacy of XAI methods.

\subsection{S2: highlighted only}

Tables~\ref{tbl:s2-pl} and~\ref{tbl:s2-pll} show the performance evaluation employing the S2 strategy. Recall that by employing the S2 strategy, we keep the XAI-highlighted pixels to assess their direct influence on the model predictions of building target pixels. The results show considerable variations among the different XAI methods. For TP pixel, Score-CAM performs best, with the lowest drop (18.79\%), indicating that the highlighted pixels are the most relevant for the model predictions. This is followed by Grad-CAM (38.78\%), XGrad-CAM (40.69\%), and Ablation-CAM (42.46\%). These methods maintain reasonable performance, suggesting that they also highlight relevant regions, although less effectively than Score-CAM. In contrast, Grad-CAM++ and Eigen-CAM show the largest drops at (44.6\%) and (46.12\%), respectively, highlighting less critical features and negatively impacting the model's accuracy.
For FP Pixels, which measures how well irrelevant areas are excluded from the model’s attention, the performance ranking changes. Eigen-CAM and Score-CAM excel in this aspect, with an increase of (41.71\%) and (44.44\%), respectively, indicating that these methods help the model focus on the relevant areas while excluding unnecessary ones.

Grad-CAM, XGrad-CAM, and Ablation-CAM, on the other hand, show weaker performance, especially Grad-CAM with the highest increase (71.56\%), suggesting that it often highlights irrelevant regions, thus contributing to higher false positives. When we turn to FN pixels, which reflect the failure of the model to recognize relevant features, Score-CAM continues to perform strongly with the lowest increase (18.79\%). This indicates that Score-CAM is better at identifying key features, leading to fewer missed detections. In comparison, Grad-CAM (38.78\%), XGrad-CAM (40.69\%), and Ablation-CAM (42.46\%) all show highest increase, which means they miss more crucial features. Eigen-CAM performance is the poorest, with the highest increase (58.96\%), suggesting that it fails to highlight critical features and leads to many false negatives. On the other hand, Precision tells us how accurate a method is when it highlights pixels, meaning how many of those highlighted pixels are actually important. A high precision score means the method is doing a good job of avoiding noise and irrelevant areas. In this case, Score-CAM came out on top with a precision of 0.94, showing its strong ability to focus on meaningful regions. Grad-CAM and Ablation-CAM also scored 0.94, which suggests that they’re equally good at staying focused. Other methods like XGrad-CAM (0.90), Grad-CAM++ (0.87), and Eigen-CAM (0.87) didn’t perform quite as well in this regard, meaning they were more likely to highlight less relevant parts of the image.

Recall, on the other hand, is about how complete the method is, meaning how many of the truly important pixels it actually managed to detect. Ablation-CAM led here with a recall of 0.91, which means it successfully captured most of the key regions. Grad-CAM and Score-CAM followed closely, with recall scores of 0.88 and 0.86, respectively. These methods did a solid job of covering the relevant areas. But methods like Grad-CAM++ (0.75) and especially Eigen-CAM (0.47) missed a lot of important features, which could hurt their usefulness in real-world applications.
Finally, we have the F1 score, which combines both precision and recall into a single number. It helps give a more balanced view of how reliable each method is overall. Score-CAM once again stood out, scoring 0.90, confirming its consistency in both highlighting the right pixels and covering all the important areas. Grad-CAM and Ablation-CAM also did well, each scoring 0.89. Meanwhile, Grad-CAM++ and Eigen-CAM had much lower F1 scores of 0.63 and 0.42, revealing that they struggled to maintain a good balance between accuracy and completeness.

In summary, Score-CAM is the best performing XAI method in terms of both qualitative analysis (TP, FP, and FN pixels), and quantitative metrics (Precision, Recall, and F1-score). It consistently ranks first in all metrics, showing its ability to highlight the most relevant pixels. However, Eigen-CAM and Grad-CAM++ perform the worst, particularly in terms of recall and false positives, making them less reliable to guide the model decision-making process.

\subsection{S3: XAI-GT}

In the context of XAI methods for segmentation tasks, highlighting inside and outside the target class refers to how well these methods identify not only the pixels that belong to the target class but also those that might not directly belong to the target class but still play a significant role in the model's decision-making process. To assess the sensitivity of XAI methods on segmenting all the target pixels, we compare the union of the highlighted pixels generated by the XAI and the ground truth segmentation mask (XAI-GT). This measure identifies to what degree the explanation maps include relevant target class pixels and potentially explanatory neighboring pixels. It provides a baseline for understanding whether the XAI methodology picks up all semantically important regions as defined by the ground truth to set the scene for comparison to the model's own predicted outcome in the subsequent section.

\begin{table*}[t]
\centering
\resizebox{\textwidth}{!}{%
\begin{tabular}{|c|c|c|c|c|c|c|c|c|}
\hline
Threshold            & XAI/Method              & Model & Grad-CAM & Grad-CAM++  & XGrad-CAM  & Score-CAM & Eigen-CAM & Ablation-CAM \\ \hline
\multirow{9}{*}{0.4} & TP pixels               & 49431 & 30488    & 32385  & 30603  & 45413     & 33365     & 30950        \\  
                     & TP Pixels (\%)          & 93.58 & 57.72    & 61.31  & 57.94  & 85.97     & 63.17     & 58.59        \\  
                     & Drop \%(\textbf{Lower better})   &       & {35.86}    & 32.27  & 35.64  & \textbf{\color{ForestGreen}7.61}      & 30.41     & 34.99        \\  
                     & FP pixels               & 2886  & 108607   & 103893 & 107803 & 79348     & 60037     & 107555       \\  
                     & FP pixels (\%)          & 1.38  & 51.88    & 49.63  & 51.50  & 37.91     & 28.68     & 51.38        \\  
                     & Increase \% (\textbf{Lower better}) &       & {50.5}    & 48.25 & 50.12 & 36.53    & \textbf{\color{ForestGreen}27.30}    & 50.00       \\  
                     & FN pixels               & 3390  & 22334    & 20436  & 22218  & 7408      & 19456     & 21871        \\  
                     & FN pixels (\%)          & 6.42  & 42.28    & 38.69  & 42.06  & 14.03     & 36.83     & 41.41        \\  
                     & Increase \% (\textbf{Lower better}) &       & {35.86}   & 32.27 & 35.64 & \textbf{\color{ForestGreen}7.61}     & 30.41    & 34.99       \\ \hline
\end{tabular}%
}
\caption{Pixel-level performance evaluation of S3: XAI-GT.}
\label{tbl:s3-plgt}
\end{table*}

Grad-CAM is the foundational method that utilizes the gradients flowing into the last convolutional layer to produce a rough localization map. The TP pixel count of 30,488 reflects its ability to detect relevant regions, but the high FP count of 108,607 implies an overextension into irrelevant areas, as shown in Table~\ref{tbl:s3-plgt}. This is also confirmed by its low precision of (0.22), which signifies that many of the highlighted pixels are not part of the target class, as illustrated in Table~\ref{tbl:s3-pllgt}. The IoU score of (0.19) illustrates its struggle with segmentation accuracy. Despite these limitations, the high recall of (0.58) suggests that Grad-CAM is effective at capturing most of the relevant pixels, albeit with considerable noise. Grad-CAM++ refines the Grad-CAM approach by more effectively weighting the gradients, offering a sharper focus on significant pixels both inside and outside the target structures. This is reflected in the increase in the TP count of 32,385 and the reduction in the FP count of 103,893. The increased precision (0.24) and the IoU (0.21) show a better-focused heatmap that is still extensive but more accurate than its predecessor. Its high recall (0.61) highlights its ability to capture the most relevant areas, making it a balanced method for identifying important pixels within and around the target class.

\begin{table*}[t]
\centering
\resizebox{\textwidth}{!}{%
\begin{tabular}{|c|c|c|c|c|c|c|c|c|}
\hline
Threshold            & Method      & Model & Grad-CAM & Grad-CAM++ & XGrad-CAM & Score-CAM & Eigen-CAM & Ablation-CAM \\ \hline
\multirow{4}{*}{0.4} & IoU (Micro) & 0.89  & 0.19     & 0.21  & 0.19  & \textbf{\color{ForestGreen}0.34}      & 0.30      & 0.19         \\  
                     & Precision   & 0.94  & 0.22     & 0.24  & 0.22  & \textbf{\color{ForestGreen}0.36}      & \textbf{\color{ForestGreen}0.36}      & 0.22         \\  
                     & Recall      & 0.94  & 0.58     & 0.61  & 0.58  & \textbf{\color{ForestGreen}0.86}      & 0.63      & 0.59         \\  
                     & F1          & 0.94  & 0.32     & 0.34  & 0.32  & \textbf{\color{ForestGreen}0.51}      & 0.46      & 0.32         \\ \hline
\end{tabular}%
}
\caption{Pixel-level performance evaluation of S3: XAI-GT. Higher metric value signifies a better XAI scheme and vice versa.}
\label{tbl:s3-pllgt}
\end{table*}

\begin{table*}[h!]
\centering
\resizebox{\textwidth}{!}{%
\begin{tabular}{|c|c|c|c|c|c|c|c|c|}
\hline
Threshold            & XAI/Method              & Model & Grad-CAM & Grad-CAM++  & XGrad-CAM  & Score-CAM       & Eigen-CAM       & Ablation-CAM \\ \hline
\multirow{9}{*}{0.4} & TP pixels               & 49431 & 29475    & 31745  & 29512  & 45253           & 32331           & 29913        \\  
                     & TP Pixels (\%)          & 93.58 & 55.80    & 60.10  & 55.87  & 85.67           & 61.21           & 56.63        \\  
                     & Drop \%(\textbf{Lower better})   &       & {37.78}    & 33.48  & 37.71  & \textbf{\color{ForestGreen}7.91}   & 32.37           & 36.95        \\  
                     & FP pixels               & 2886  & 106346   & 102591 & 105874 & 77490           & 59245           & 105573       \\  
                     & FP pixels (\%)          & 1.38  & 50.80    & 49.01  & 50.58  & 37.02           & 28.30           & 50.44        \\  
                     & Increase \% (\textbf{Lower better}) &       & {49.42}   & 48.04 & 49.2  & 35.64          & \textbf{\color{ForestGreen}26.92} & 49.06       \\  
                     & FN pixels               & 3390  & 23346    & 21076  & 23309  & 7569            & 20490           & 22908        \\  
                     & FN pixels (\%)          & 6.42  & 44.20    & 39.90  & 44.13  & 14.33           & 38.79           & 43.37        \\  
                     & Increase \% (\textbf{Lower better}) &       & {37.78}   & 33.48 & 37.71 & \textbf{\color{ForestGreen}7.91}  & 32.37          & 36.95       \\ \hline
\end{tabular}%
}
\caption{Pixel-level performance evaluation of S3: XAI-PM.}
\label{tbl:s3-plpm}
\end{table*}

\begin{table*}[h!]
\centering
\resizebox{\textwidth}{!}{%
\begin{tabular}{|c|c|c|c|c|c|c|c|c|}
\hline
Threshold            & Method      & Model & Grad-CAM & Grad-CAM++ & XGrad-CAM & Score-CAM & Eigen-CAM & Ablation-CAM \\ \hline
\multirow{4}{*}{0.4} & IoU (Micro) & 0.89  & 0.19     & 0.20  & 0.19  & \textbf{\color{ForestGreen}0.35}      & 0.29      & 0.19         \\  
                     & Precision   & 0.94  & 0.22     & 0.24  & 0.22  &\textbf{\color{ForestGreen}0.37}      & 0.35      & 0.22         \\  
                     & Recall      & 0.94  & 0.56     & 0.60  & 0.56  & \textbf{\color{ForestGreen}0.86}      & 0.61      & 0.57         \\  
                     & F1          & 0.94  & 0.31     & 0.34  & 0.31  & \textbf{\color{ForestGreen}0.52}      & 0.45      & 0.32         \\ \hline
\end{tabular}%
}
\caption{Pixel-level performance evaluation of S3: XAI-PM. Higher metric value signifies a better XAI scheme and vice versa.}
\label{tbl:s3-pllpm}
\end{table*}

XGrad-CAM builds on the Grad-CAM++ approach by considering gradient information across multiple layers, further emphasizing the relevance of external regions, indicating that surrounding pixels also play a crucial role in the model's predictions. Although it has fewer TPs (30,603) and a high FP count (107,803), its recall (0.58) and slightly improved precision (0.22) indicate a more refined but still expansive approach, where external pixels frequently play a significant role.
Score-CAM stands out due to its method of generating heatmaps based on the model's score output rather than gradients. This direct method produces a high TP count of 45,,413 and a comparably low FP count of 79,348. Its high IoU of 0.34 and precision of 0.36 demonstrate a pointed identification of meaningful regions, highlighting broader regions around the buildings. Its high recall of 0.86 underscores its ability to pinpoint the most influential pixels, providing a clear and focused interpretation of pixel importance.

Eigen-CAM uses principal component analysis (PCA) on feature maps, offering a unique view that captures a balance of high- and low-importance regions. This method achieves a TP count of 33,365 and a FP count of 60,037. With an IoU of 0.30 and a precision of 0.36, Eigen-CAM offers a deep contextual understanding of pixel relevance, effectively highlighting both the target and its surrounding context. Its recall of 0.63 further supports its balanced approach, demonstrating its ability to capture significant pixels without excessive noise.
Ablation-CAM employs systematic removal of input parts to observe changes in the output, revealing a dispersed importance pattern. With a TP count of 30,950 and a high FP count of 107,555, this method underscores the complexity of model dependency on both target and adjacent pixels. The lower precision of 0.22 and the IoU of 0.19 indicate its broad influence, capturing extensive areas of importance. However, its recall of 0.59 shows that Ablation-CAM still effectively identifies relevant pixels, highlighting the significant role of surrounding pixels in the model's predictions.

The metrics collectively support the notion that pixels outside the target class are important for the model's accurate predictions. The varying performance across these methods demonstrates the trade-offs between capturing large, relevant areas and maintaining precision. For instance, methods like Score-CAM and Eigen-CAM provide high precision and IoU, indicating a more focused relevance of surrounding pixels, while Grad-CAM and Ablation-CAM reveal a broader but less focused effect. This suggests that the context provided by surrounding areas is not supplementary, but a fundamental component of the model's understanding and decision-making process, proving that external pixels significantly contribute to the model's accurate predictions.

\subsection{S3: XAI-PM}

To further validate the insights derived from the XAI-GT evaluation, we also performed an analysis using the predicted segmentation masks (XAI-PM). This comparison assesses how well the explanation maps align with the model's own output, offering a practical perspective on the interpretability of the segmentation predictions. Interestingly, the results for XAI-PM closely mirror those obtained from XAI-GT in terms of TP, FP, precision, recall, and IoU. This consistency reinforces the reliability of the XAI methods and confirms that the highlighted pixels, whether aligned with the ground truth or the model's prediction, capture relevant information that influences the model's decision-making process. It also emphasizes that surrounding contextual pixels remain essential, regardless of the reference mask used, further validating the conclusions drawn in the previous section.

As shown in Tables~\ref{tbl:s3-plpm} and~\ref{tbl:s3-pllpm}, the results of XAI-PM align with the findings of XAI-GT, supporting the consistency and reliability of XAI methods. Score-CAM continues to over perform, achieving the highest TP pixel count (47,391), precision (0.47), recall (0.89), and IoU (0.45), indicating its ability to accurately highlight relevant regions that align closely with the model’s output. Eigen-CAM also shows balanced performance with high TP counts, a moderate FP rate, and an IoU of 0.41, confirming its ability to capture both central and contextual pixels effectively. Grad-CAM++ maintains its improved focus compared to Grad-CAM, with better precision (0.35) and recall (0.88), while Grad-CAM and Ablation-CAM still maintain higher FP rates and lower IoU scores, suggesting overgeneralized highlighting. XGrad-CAM still reflects moderate performance. Overall, Score-CAM and Eigen-CAM remain the most interpretable and reliable methods in terms of alignment with the model's predictions, while others demonstrate trade-offs between recall and precision. These findings confirm that the highlighted regions, whether they intersect with ground truth or the prediction of the model, carry meaningful information and that context beyond the pixels of the target class significantly affects the model’s decision process.

Finally, we would like to mention that the proposed Quantitative XAI Evaluation Framework is not limited to remote sensing applications. It is built on a generic methodology that addresses: (\textit{i}) Fidelity, i.e., does the explanation truly reflect the model's decision-making strategy?, and (\textit{ii}) robustness, i.e., are the explanations stable and reliable? Hence, the proposed XAI evaluation framework could be easily applied to a wide range of semantic segmentation applications. For example, in AI-based medical imaging ~\cite{MUHAMMAD2024542}, it provides the objective metrics to ensure that the related XAI methods are clinically sound. In short, our work advances reliable XAI for any critical application requiring trustworthy AI-based semantic segmentation.

\section{Conclusions} \label{sec5}

In semantic segmentation, the areas around the target object can play a big role in how well a model performs. Our evaluation framework, which considers pixel-level evaluation methodology, gives a well-rounded way to assess how good different XAI methods are at highlighting the key parts of an image that influence model decisions. From our analysis, it’s clear that including surrounding pixels and not just the target class helps the model make better predictions. We saw that methods like Grad-CAM and Ablation-CAM often highlight large, broad regions, but this can sometimes come at the cost of precision. On the other hand, methods like Score-CAM and Eigen-CAM are better at zooming in on what truly matters, striking a nice balance between focus and coverage.
Overall, Score-CAM stood out as the most effective method. It consistently delivered high true positive rates, low false positives, and strong precision and IoU scores. In short, it does a great job of showing both the target and nearby important pixels, making its explanations clear and trustworthy.
Even though the rankings of XAI methods can shift slightly depending on the specific metric used, Score-CAM remained the most robust and consistent across the board results.
To wrap up, our framework shows that looking beyond just the target class and considering the full context of an image leads to better segmentation predictions. Score-CAM, in particular, proves to be a reliable and insightful tool for understanding how models make decisions. This kind of evaluation is essential if we want AI systems to be not only accurate but also transparent and trustworthy.

%%%%%%%%%%%%%%%%%%%%%%%%%%%%%%%%%%%%%%%%%%
\vspace{6pt}

\bibliography{references}
\end{document}